\documentclass[a4paper]{article}

\usepackage{INTERSPEECH2022}

\usepackage{mathtools}
\usepackage{multicol}
\usepackage{hyperref}
\usepackage[table,xcdraw]{xcolor}
\usepackage{amsmath,graphicx}
\DeclarePairedDelimiter\ceil{\lceil}{\rceil}
\DeclarePairedDelimiter\floor{\lfloor}{\rfloor}

\title{Speech Sequence Embeddings\\ using Nearest Neighbors Contrastive Learning}
\name{Robin Algayres$^{1,2}$, Adel Nabli$^{1,2}$, \textit{Benoît Sagot}$^2$, \textit{Emmanuel Dupoux}$^{1,2,3}$}
\address{$^1$ENS-PSL, EHESS, CNRS, Paris
  $^2$Inria, Paris, $^3$Facebook AI Research}
\email{robin.algayres@inria.fr, adel.nabli@umontreal.ca, benoit.sagot@inria.fr,
emmanuel.dupoux@inria.fr}

\begin{document}

\maketitle
\vspace{-2.em}
\begin{abstract}
We introduce a simple neural encoder architecture that can be trained using an unsupervised contrastive learning objective which gets its positive samples from data-augmented k-Nearest Neighbors search. We show that when built on top of recent self-supervised audio representations \cite{hsu21hubert,baevski20w2v2,cpc_morgane}, this method can be applied iteratively and yield competitive SSE as evaluated on two tasks: query-by-example of random sequences of speech, and spoken term discovery. On both tasks our method pushes the state-of-the-art by a significant margin across 5 different languages. Finally, we establish a benchmark on a query-by-example task on the LibriSpeech dataset to monitor future improvements in the field\footnote{Code is available at \href{https://gitlab.cognitive-ml.fr/ralgayres/speechseqembeddings}{https://gitlab.cognitive-ml.fr/ralgayres/speechseqembeddings}}. 
\end{abstract}

\section{Introduction}
\label{sec:intro}
Recently, research into self-supervised representations for speech \cite{oord18cpc,cpc_morgane,baevski20w2v2,hsu21hubert}
has made impressive progress, resulting in audio features that are much better than the traditional MFCC features in representing speaker invariant phonetic content. The learned representations typically span a rather short and fixed-length time window (10 or 20ms) and can be fine-tuned for ASR with few or even no labels \cite{baevski20w2v2,baevski21unsup_w2v}, or used after clustering as proxy phonemes for language modelling or speech synthesis \cite{lakhotia21gslm,polyak21resynt}.
However, such representations are still not good enough for other downstream tasks that need to handle larger word-like units that typically span larger time windows (100ms-1000ms) and have varying duration depending on speech rate. These tasks include word segmentation \cite{eskmeans,bhati2021scpc}
term discovery \cite{jansenzs,rasanen,rasanen2,rasanen3,lyzinski,bhati} and unsupervised alignment of text and speech \cite{baevski21unsup_w2v,glass2018unsup_align}. Despite recent research efforts into developing word embeddings also known as speech sequence embeddings (SSE), \cite{last20caesiamese,nils18wordemb,settle2016siamese,chung2016audio,badino2014wordemb}
their performances are low and difficult to compare for lack of a common evaluation benchmark. 

\begin{figure*}
 \centering
  \includegraphics[width=\textwidth]{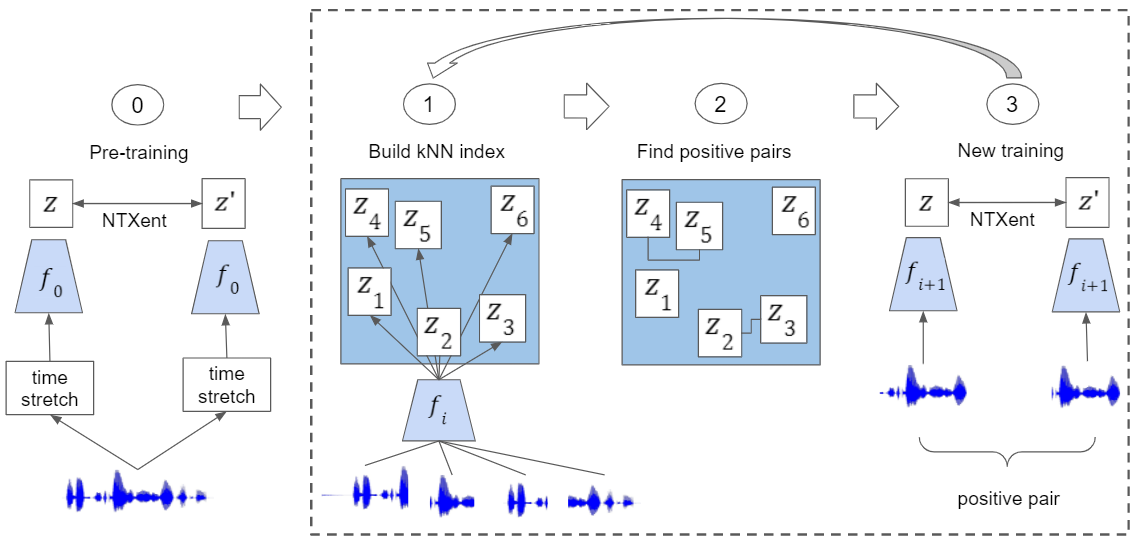}
  \caption{The main steps of our method. 0) Train a model $f_0$ with the NTXent loss on time-stretched pairs of sequences of speech. 1) Use $f_0$ to encode many random sequences of speech from the corpus. 2) List close embeddings in the kNN as positive pairs. 3) Use the speech sequences associated to the positive pairs to train a new model $f_1$. Go back to step 1 using $f_1$ instead of $f_0$ and iterate.}
  \label{main_algo}
\end{figure*}

Our contribution is twofold. First, we introduce a benchmark for SSE based on a query-by-example task built on LibriSpeech \cite{librispeech}. It uses the Mean Average Precision (MAP) metric computed over random query phoneme-ngrams. Previous work have shown that MAP is a good proxy for word segmentation \cite{carlin11map,algayres20evalemb}. We show that MAP based on phoneme-ngrams is preferable to a MAP on words, which is too easy and does not reflect the nature of the downstream tasks in unsegmented continuous speech.\\
Second, we introduce a new method for learning SSE based on a combination of ideas from representation learning. The model is a convolution and transformer-based embedder trained with the NTXEnt contrastive loss \cite{ntxent}.
Building on similar ideas in vision and speech, we select our positive examples through a mix of time-stretching data augmentation \cite{chen20simclr}
and k-Nearerst Neighbors search \cite{dwidebi21googleknn,thual18knn}. Figure \ref{main_algo} gives an overview of our method. To evaluate our method, we test our model on 5 types of acoustic features: MFCCs, CPC \cite{oord18cpc,cpc_morgane} 
HuBERT \cite{hsu21hubert} and Wav2Vec 2.0 (Base and Large) \cite{baevski20w2v2}.  We pick the best method from our LibriSpeech benchmark and show that when applied without any change to the task of spoken term discovery as defined in the zero resource challenges \cite{zs20}, we beat the state of the art on the NED/COV metric by a large margin in 5 new datasets. 
\vspace{-1.em}
\section{Related work}\label{related}
\textbf{Speech Sequence Embeddings.} One straightforward method to obtain an SSE is to perform a max pooling along the time axis of the frame-wise speech features. This method (combined with a PCA) have been used in \cite{baevski21unsup_w2v} to represent speech sequences obtained by unsupervised phoneme segmentation. Another method involves downsampling the sequence into a fixed number of frames and concatenate them into a single vector \cite{nils18wordemb}. Other methods use bottleneck auto-encoders \cite{nils18wordemb}.
More advanced methods use Siamese loss \cite{settle2016siamese} or contrastive learning \cite{herman2021infonce} and rely on unsupervised ways to obtain positive pairs. Similarly, auto-encoders with a correspondence objective \cite{kamper2015cae} can encode a member of a pair and reconstruct the other member of the pair. Correspondance and Siamese losses can be combined effectively in a model called CAE-Siamese that is on par with the current state-of-the-art in SSE models \cite{last20caesiamese,algayres20evalemb}.\\
At present, there is no common corpus and evaluation metric to evaluate the intrinsic quality of SSE models. For instance, \cite{chung2016audio,herman2021infonce,settle2016siamese,algayres20evalemb}
use pre-segmented word tokens to perform a query-by-example task measured by MAP \cite{carlin11map}. Others rely on an ABX test on word triplets \cite{nils18wordemb}.
or word classification \cite{badino2014wordemb} 
All these research papers compute their metrics on different corpora featuring different languages. This plurality of metrics and corpora makes it hard to monitor improvement in the field. We propose to create a common benchmark using one of the most popular metric so far: the MAP, on a widely known corpus: the LibriSpeech dataset \cite{librispeech}.\\
\textbf{Spoken term discovery.} This task consists in searching audio recordings for speech sequences that are similar in acoustic space. One popular method is to use Segmental DTW \cite{seg_dtw} on hand-crafted features, mainly MFCC and PLP, to find subparts of two utterances that have a good DTW alignment score. The main problem with this method, as shown by \cite{kamper2021}, is that for some unknown reason, even though it works well on traditional hand-crafted speech features like MFCC and PLP features, it fails to benefit from the theoretically higher quality frame-level self-supervised embeddings like Wav2vec 2.0, CPC or Hubert. Another method which does not share this limitation, has been proposed by \cite{thual18knn}. It works by encoding all speech sequences of a corpus of a given length range with an SSE and look for pairs with low cosine distance using accelerated k-Nearest Neighbors search. We adopt this method here, and the NED/COV metric from the Zerospeech challenges \cite{zs20} developped in \cite{ludusan2014bridging} 
to evaluate spoken term discovery.\\




\vspace{-1em}
\section{Method}\label{sse}




\subsection{Contrastive learning}\label{ntxloss}

The NTXent is a contrastive loss \cite{ntxent} that takes as input a batch $P^+$ of positive pairs of embedded speech sequences $P^+=((z_0,z_0^+),(z_1,z_1^+),...,(z_n,z_n^+))$. For each item taken from a positive pair,  all remaining $2(n-1)$ items from the other pairs are used to form negative pairs. This loss minimizes the distance within the positive pairs while maximizing the distance within the negative pairs. For each item $z_i \in P^+$, the loss function is written:\\
\begin{equation}
L_{nce}(z_i,P^+)=-log(\frac{\exp(sim(z_i,z_i^+)/\tau)}{\sum_{j\le2n j\ne i}\exp(sim(z_i,z_j)/\tau)})
\label{moneq}
\end{equation}

where $\tau$ is the softmax temperature parameter and $sim$ is the cosine similarity.

\subsection{Stretch-invariant pretraining}\label{initial}

At initialisation (step 0 in Figure \ref{main_algo}), our SSE model is pretrained with the NTXEnt loss \cite{ntxent}
on batches of positive pairs obtained by data augmentation. First, we apply Voice Activity Detection (rVAD \cite{rvad}) to segment LibriSpeech into continuous Vocal Activity (VA) segments without any silence. Next, we use data augmentation on these speech sequences.
using WavAugment \cite{kharitonov21dataaug}. 
Preliminary experiments showed that of the various augmentations (pitch, reverberation, additive noise, masking, etc), only time stretch had a large and systematic effect. This makes sense as variations in speech rate is a cause of a major change in the acoustic representation of a given speech sequence: its length. We therefore use time stretch as the sole data augmentation method.

We create a set of positive pairs from a single sampled VA segment, $v$ of duration $d$ as follows. From $v$, we create two variants by stretching its duration by a random factor in $[0.5,1.8]$, giving rise to $v_1$ and $v_2$ with respective durations $d_1$ and $d_2$. These variants are converted into sequences of frame-level features, $f_1$ and $f_2$, using an off-the-self acoustic features encoder with frame rate $r$ ($r=50$ or $r=100$ frames per second depending on the encoder). We create positive pairs by sampling sub-sequences of frames in $f_1$ and $f_2$ that can be realigned to the same original acoustic signal from $v$.
Specifically, we sample in $f_1$ a starting frame at index $s$ and ending frame at index $e$ so that $r*0.08 \le e-s \le r$. We do not want to sample speech sequences that are shorter than one phoneme ($80ms$) or longer than most words ($1s$). Then the corresponding sub-sequence in $f_2$ is found between the frames at indices $\floor*{s\frac{d_2}{d_1}}$ and $\ceil*{e\frac{d_2}{d_1}}$. We iterate this process until the number of positive pairs sampled this way reaches the batch size. The negative pairs are formed by recombining items from different positive pairs. Therefore, it is important that the sequences sampled from $f_1$ are temporally shifted by at least one phoneme length ($80ms$) so that they correspond to distinct phonetic contents. To ensure that, when sampling a starting frame index $s$ in $f_1$, we ensure that $s\%(0.08*r)=0$. The same procedure is applied for the end index $e$.

\subsection{Self-labelling with k-NN search}\label{loopknn}

We apply the method described in \cite{thual18knn} to find more positive pairs with k-Neareast-Neighbors search (step 1 and 2 in Figure \ref{main_algo}). 
We extract sub-sequences of frames within each VA ensuring they are all shifted by the average phoneme length ($80ms$). Next, we apply the SSE model pretrained according to our initialisation method (section \ref{initial}) to turn sampled sub-sequences of frames into embeddings, which are indexed for k-NN search using the FAISS library \cite{faiss}. 
For each embedding $s$, the search returns $N$ closest neighbors according to the cosine distance. We trim down this number to construct pairs that are more likely to have the same phonetic content as follows: 
 First, all neighbors that are temporally overlapping with $s$ are removed. Second, when two neighbors are temporally overlapping with each other, we keep only the one that has the smallest cosine distance with $s$ (Non-Maximal-Suppression~\cite{thual18knn}). 
 Finally, we apply a distance threshold above which all pairs are discarded. We set it such that half of speech sequences indexed in the k-NN graph have at least one selected pair. This was estimated on a sample of LibriSpeech (which functions as our dev set) where we found out that the proportion of speech sequence (excluding silences) with a unique transcription was around $50\%$. All selected pairs were used to form batches of positive pairs to retrain a new SSE model from scratch (step 3 in Figure \ref{main_algo}). We iterate the procedure by re-embedding the speech sequences into a new k-NN index, re-sampling positive pairs and retraining another SSE model. 

\begin{figure*}[t!]
        \centering
        \includegraphics[width=.2\linewidth]{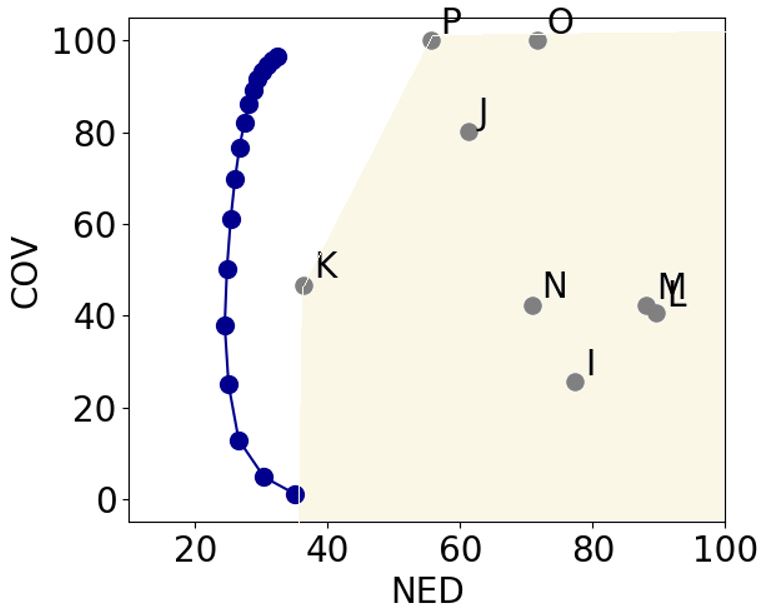}\hfill
        \includegraphics[width=.2\textwidth]{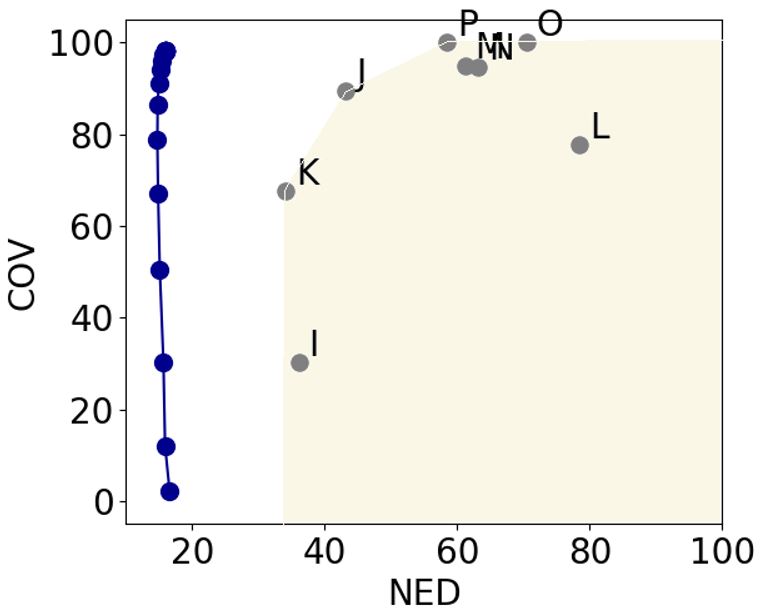}\hfill
        \includegraphics[width=.2\textwidth]{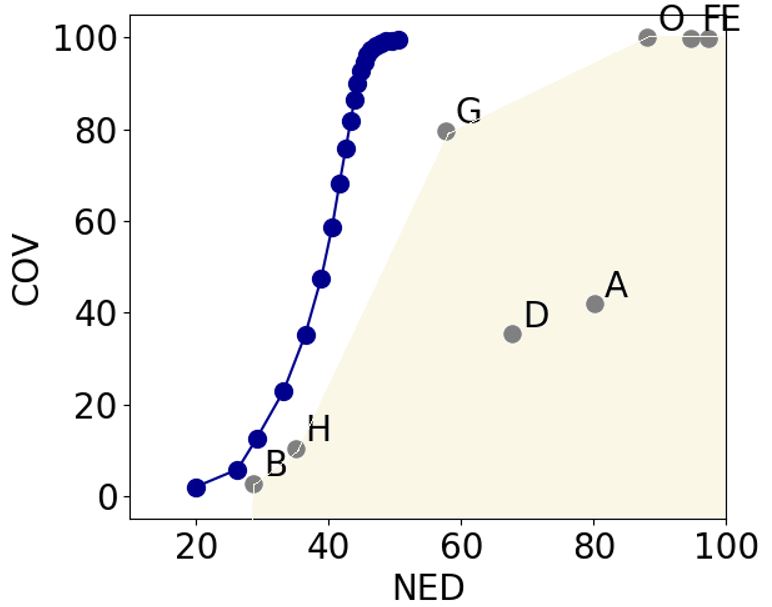}\hfill
        \includegraphics[width=.2\textwidth]{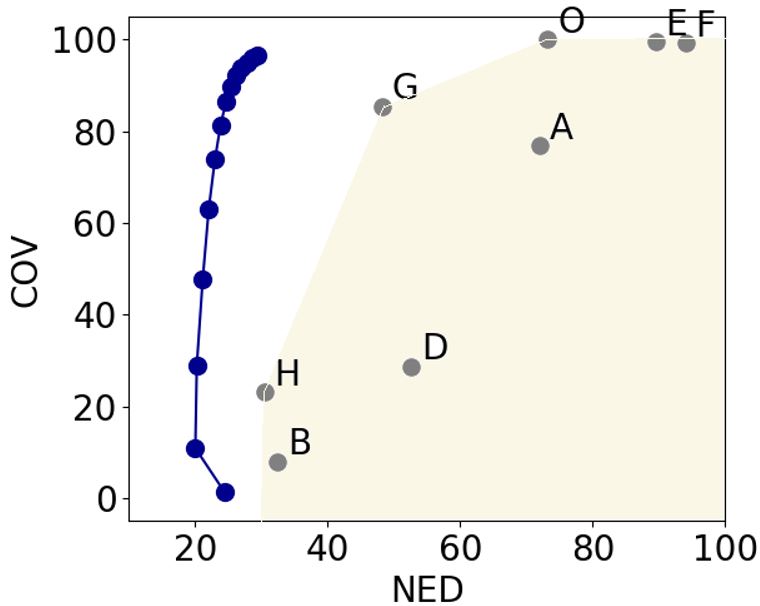}\hfill
        \includegraphics[width=.2\textwidth]{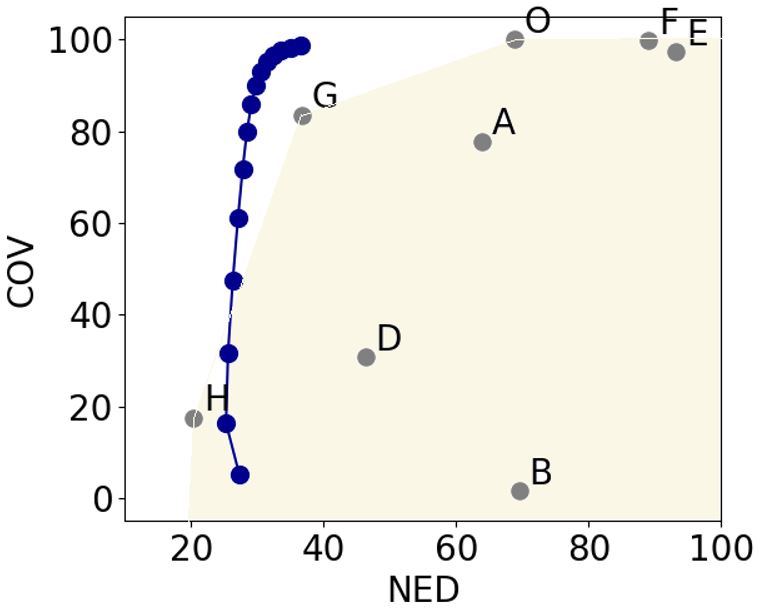}\hfill
(a) Buckeye \quad \quad \quad \quad \quad(b) Xitsonga \quad \quad \quad \quad \quad(c) Mandarin \quad \quad \quad \quad \quad (d) English \quad \quad \quad \quad \quad(e) French
        \caption{NED/COV curves on Zerospeech corpora. Our method is represented with a line of blue dots and each other competitors as grey points: Garcia-Granada (A \cite{garcia}), Jansen (B,\cite{jansenzs}),  Räsänen (L,M,N \cite{rasanen} and D \cite{rasanen2} and G,H \cite{rasanen3}) , Kamper (O \cite{eskmeans} and P    \cite{besgmm}), Lynsinski (I,L,J \cite{lyzinski}), Bhati (E,F\cite{bhati}).}
\label{nedcov}
\end{figure*}

\subsection{Evaluation
 metrics}\label{evaluation}

We evaluated the SSE models on the basis of two tasks: a query-by-example task measured by MAP and a word discovery task measured by NED/COV. \\
\textbf{Query-by-example}. Given a query (a speech sequence), the task is to retrieve all sequences with the same transcription in an audio file. 
Previous works have used lists of words extracted from an audio file using forced alignment\cite{carlin11map}. This task is a little too easy, since the words are disjoint intervals in the speech signal, while a retrieval task on an unsegmented audio file would have to deal with potentially overlapping speech sequences that differ from the query by only one phoneme. To push the task closer to an unsegmented use case, here, we compute MAPs over randomly selected phoneme-ngrams, i.e non-silent speech sequences that start and end at phoneme boundaries. To avoid long computation of MAP, we reduce the number of such sequences by picking only those that are shorter than 1 second. Most of them will be phoneme-ngrams with unique transcription in force-alignment. We also remove all phoneme-ngrams with unique transcriptions. Prior experiments showed that it does not change the MAP scores much but drastically reduces computation time.
Given a query, we can compute precision and recall of the retrieved sequences as a function of the distance threshold $r$. The MAP score is obtained by varying $r$ between 0 and infinity and integrating the precision/recall curve. It results in a score between 0 and 1 (higher is better). We compute it using the pytorch-metric-learning library \cite{pytorch-metric}.\\
\textbf{Term Discovery metric}. 
In its simplest form, spoken term discovery consist in finding clusters or pairs of speech segments that have the same transcription from unsegmented audio. It is evaluated by the NED/COV as introduced in the zero speech challenges \cite{zs20}. For each discovered pair, the Normalised Edit Distance (NED) is computed as the edit distance normalised by the length of the loangest item in the pair. The coverage (COV) is given by the proportion of phonemes in the corpus that is covered by least one speech sequence from the sets of pairs. There is typically a tradeoff between these two measures: low NED systems have typically low COV, and high COV have high NED.

\begin{figure}[t!]
    \begin{center}
    \includegraphics[width=1.05\linewidth]{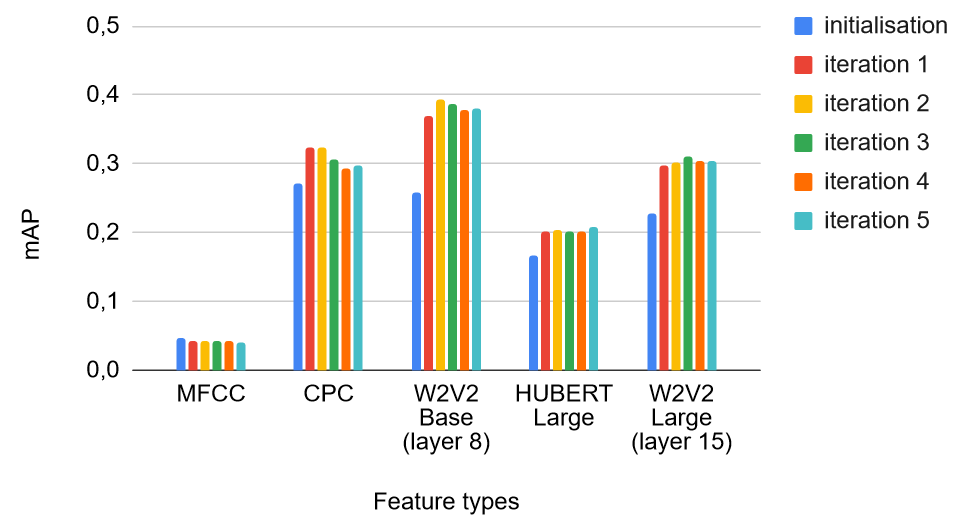}

    \caption{Phoneme-ngrams MAP computed on LibriSpeech dev-clean for different iteration of our model and different input feature types}

    \label{map_per_features}
    \end{center}
\vspace{-2em}
\end{figure}

\begin{table}[]
\centering
\resizebox{\linewidth}{!}{
\begin{tabular}{ll cccc}
\hline
Supervision  & Models             & dev-clean    & dev-other & test-clean     & test-other     \\ \hline
unsup.      & Max-pooling        & 0.07         & 0.048     & 0.07           & 0.047          \\
self-sup.   & CAE-Siamese        & 0.21                                 & 0.154                             & 0.212                                  & 0.151                                  \\
self-sup.   & Ours (iter. 2) & \textbf{0.398} & \textbf{0.307}                    & \textbf{0.399} & \textbf{0.305} \\ \hline
weakly-sup. & Topline            & 0.789                                & 0.647                             & 0.784                                  & 0.648                                  \\ \hline
\end{tabular}}

\caption{Phoneme-ngrams MAP computed on LibriSpeech held-out sets for different SSE models. All models take as input features the Wav2vec2.0 Base at layer 8}
\label{benchmark}
\vspace{-1em}
\end{table}

\begin{table}[]
\resizebox{\linewidth}{!}{
\begin{tabular}{lcccccc}
\hline
\rowcolor[HTML]{FFFFFF} 
Models                                     & buckeye        & xitsonga                      & mandarin                      & french                        & english                       & \textit{average} \\ \hline
Max-pooling        & 0.05           & 0.053 & 0.075 & 0.039 & 0.052 & 0.054           \\
CAE-Siamese        & 0.16           & 0.23                          & 0.26                          & 0.2                           & 0.19                          & 0.208            \\
Ours (iter. 2) & \textbf{0.235} & \textbf{0.362}                & \bf 0.277               & \textbf{0.283}                & \textbf{0.346}                & \textbf{0.301}   \\ \hline
Topline            & 0.751          & 0.948                         & 0.822                         & 0.71                          & 0.857                         & \textit{0,818}            \\ \hline
\end{tabular}}
\caption{Phoneme-ngrams MAP computed on Zerospeech corpora for different SSE models. All models take as input features the Wav2vec2.0 Base at layer 8}
\label{map_per_corpus}
\vspace{-3em}
\end{table}

\subsection{Datasets}

For training and evaluation of our models we used six different speech datasets that have been force-aligned with a phonetic transcription. Models are trained on the speech only and phonetic force-alignments are used for evaluation metrics.
First, we use the LibriSpeech dataset \cite{librispeech} composed of a 960 hours training set and four 5 hours held-out datasets for development and test called: dev-clean, base at test-clean and test-other. Then, to study generalization, we used five speech corpora from the Zerospeech challenges \cite{zs20}: English (45hours), Xitsonga (2h30), Mandarin (2h30), French (24h) and the Buckeye (5 hours of conversational English). As these corpora do not have a held-out test set, evaluation is performed using the training set, a common practice in unsupervised speech processing \cite{zs20}.

To compute a MAP score on the LibriSpeech dataset, we sample phoneme-ngrams from the four held-out datasets as described in section \ref{evaluation}. It results in around 700,000 phoneme-ngrams in each held-out dataset. On the Zerospeech corpora, we hypothesise a test-set by sampling from the training set 150,000 non unique phoneme-ngrams.

\subsection{Models architectures}

Our SSE model ($f$ in Figure \ref{main_algo}) is a neural network which starts with a layer-normalisation and a 1-dimensional convolution layer with 512 channels (kernel size 4, stride 1) followed by Gated-linear-units and drop out. Sinusoidal Positional embeddings are added after the convolution layer. After going through a transformer layer with 4 attention heads and 512 units, the resulting features are max-pooled along the time axis to form a fixed-size embedding of the input speech sequence. During training, we add a projection head composed of two 512 neurons linear layers and a ReLU activation in between the layers. The model is trained by back-propagation with Adam optimiser and 0.0001 initial learning rate. The temperature parameter of the NTXEnt contrastive loss is set to 0.15. 

We feed our SSE model on 4 frame-level pre-trained features: CPC (last layer as in \cite{cpc_morgane}), Wav2vec2.0 (Base at layer 8 and Large at layer 15 as in \cite{baevski21unsup_w2v}), and Hubert Large (last layer as in \cite{hsu21hubert}). 
We add MFCC (40 coefficients) for comparison. All hyper-parameters were chosen by gridsearch on the LibriSpeech to maximizes the MAP scores on the dev-clean set.

We compare this model to two other methods: Max-pooling and CAE-Siamese. 
Max-pooling is a common baseline SSE model. 
It consists in max-pooling input frame-level features across time to get a fixed-size vector. 
The CAE-Siamese referenced in \cite{last20caesiamese,algayres20evalemb} is on par with the current state-of-the-art in SSE models. We took the implementation from \cite{algayres20evalemb} which is a three-layers LSTM encoder-decoder trained with a weighted average of the Siamese loss as in \cite{settle2016siamese} and the Correspondance loss from \cite{kamper2015cae}. Positive pairs for training are sampled using PLP features as in \cite{algayres20evalemb}. We know from \cite{kamper2020comparison} that the Correspondance loss benefits from recent frame-level speech representation. Therefore, we train the CAE-Siamese with Wav2vec2.0 features (Base model at layer 8) to give a general idea of the current best practices in SSE modelling.

\section{Results and discussion}

\subsection{Mean Average Precision}

Figure \ref{map_per_features} shows the MAP scores evaluated on LibriSpeech dev-clean datasets for different frame-level input features and different training steps (initialisation and k-NN iterations). Overall, our method reaches a peak of performance after two iterations of k-NN self-labelling. Across input features, the Wa2vec2.0 features from the Base model at layer 8 lead to the best MAP scores. From now on, we use this input features for all following experiments.

Table \ref{benchmark} is our benchmark across different SSE models on LibriSpeech held-out datasets. As expected, the CAE-Siamese is consistently better than the Max-pooling. Yet it is far from reaching the performance of our method after two iterations of k-NN self-labelling. 

As a topline system, we train our neural network architecture on positive pairs of speech sequences that have each the same phonetic transcription. These positive pairs are sampled using the time-aligned transcriptions of our training sets. Unsurprisingly, the MAP scores of this weakly-supervised topline are way above all others. 

To show generalisation, we chose the configuration of our SSE model that works best on our LibriSpeech benchmark and applied it directly on the five corpora of the Zerospeech challenges. In table \ref{map_per_corpus} again, Max-pooling and CAE-Siamese get much lower scores than our method as well as the weakly-supervised topline. Interestingly, the Wav2vec2.0 were pre-trained on the LibriSpeech corpus but transfer well as frozen features across corpora and languages.

Yet, some papers have warned on the reliability of MAP and ABX \cite{algayres20evalemb,abdullah2021familiar}. They show that only large improvements in those metrics produce higher performances in downstream tasks.\\

\subsection{NED/COV}


Figure \ref{nedcov} shows the NED/COV scores of our model on the Zerospeech's dataset compared to other models submitted for previous editions of the challenge. Some of these models performed additional tasks that what we do not do here such as clustering and finding word boundaries. Therefore we only run the evaluation metrics relevant to the matching task. 
Competitors' submissions to the Zerospeech challenges are represented with grey dots on figure \ref{nedcov} as they were asked to submit only one set of clusters. We do not impose this constraint on us. For each corpus, using our trained SSE model, by varying the threshold mentionned in section \ref{evaluation}, we create a range of twenty submissions per corpora which appear as a blue dots on figure \ref{nedcov}. Strong NED/COV scores are in the upper left corners. Regarding the NED/COV results, whatever the threshold that we choose for our method, we get better NED/COV than all other systems on all datasets (except one submission on the French dataset but we beat that submission on the other datasets). 

\section{Conclusion}

We introduced a new Speech Sequence Embedding model that leverages time-stretching data augmentation as well as nearest neighbors search. Across several corpora, our model shows much higher results on our ngram-based MAP LibriSpeech metric than a previous competitive model (CAE-Siamese), although there is still room for improvements based on our weak supervision experiment. We open source our model and our MAP metric in the hope it will encourage researchers to use the latter to evaluate SSE models. 
On the NED/COV metric, our model push the state-of-the-art by a large margin compared to other submissions from the Zerospeech challenges, opening up the possibility of further improvement in speech segmentation tasks.


\section{Acknowledgements}

This work was funded in part, to the authors in their academic capacities,
by the Agence Nationale pour la Recherche (ANR-17-EURE-0017 Frontcog, ANR-10-IDEX-0001-02 PSL*, ANR-19-P3IA-0001 PRAIRIE 3IA Institute), CIFAR (Learning in Machines and Brains) and Facebook AI Research (Research Grant). This work was performed using HPC resources from GENCI-IDRIS (Grant 2021-[AD011011217]).

\bibliographystyle{IEEEtran}

\bibliography{mybib.bib}
\appendix

\section{Appendix}

\subsection{Choice of Wav2vec2.0 layer}\label{sec:supp}

We provide a per-layer analysis of Wav2vec2.0 Base and Large to justify the choice of the layer used in this work. We used the ABX tasks from the ZeroSpeech Challenge 2021 \cite{nguyen2020zs21}. The task computes, for a given contrast
between two speech categories A and B (e.g., the contrast between triphones ‘aba’ and ‘apa’), the probability that two sounds belonging to the same category are closer to one another than two sounds that belong to different categories. 


After evaluating Wav2vec2.0 Base and Large with these tasks, the best layer for Wav2vec2.0 Base is at index 8 and the best layer for Wav2vec2.0 Large is at index 15 which is consistent with \cite{baevski21unsup_w2v} where the authors found that the 15th layer of Wav2vec2.0 Large is the best for phone recognition.

\subsection{Scaling to large datasets and models}\label{sec:suppb}

We scaled our method to the whole LibriSpeech corpus which did not improve the MAP scores. We also implemented a sequence-to-sequence transformer (4 layers for the encoder and 4 layers for the decoder) that takes in input our SSE built on top of Wav2vec2.0 and decodes them into the discrete HuBERT units from \cite{gslm}. The gradient is backpropagated through the sequence-to-sequence and the SSE (but not Wav2vec2.0). We trained this model on the whole LibriSpeech dataset both using an SSE pre-trained or untrained. Yet all our attempts did not provide better performances than the one described in this paper. Scaling this SSE to larger datasets and models remains an open question.

\begin{table}[]
\centering
\resizebox{0.8\linewidth}{!}{
\begin{tabular}{l cc}
\hline
layer  & abx within & abx across    \\ \hline
frontend	& 13.74\%	&22.97\% \\
layer 0	&11.17\%	&17.07\% \\
layer 1	&10.05\%	&14.94\% \\
layer 2	&9.50\%	&13.28\% \\
layer 3	&8.72\%	&11.14\%  \\
layer 4	&7.24\%	&8.15\% \\
layer 5	&7.48\%	&7.95\% \\
layer 6	&8.04\%	&8.68\% \\
layer 7	&8.11\%	&8.35\% \\
layer 8	&\bf6.92\%	&\bf7.78\% \\
layer 9	&7.70\%	&10.99\% \\
layer 10	&26.77\%&	35.55\% \\
layer 11	&25.99\%&	33.04\% \\
final proj	&26.42\%	&33.06\% \\ \hline
\end{tabular}}

\caption{Per layer ABX scores for Wav2vec2.0 Base}
\label{fig:abx_small}
\end{table}

\begin{table}[]
\centering
\resizebox{0.8\linewidth}{!}{
\begin{tabular}{l cc}
\hline
layer  & abx within & abx across    \\ \hline
frontend &12.68\%	&20.66\%	\\
 layer 0	&12.38\%	&19.38\% \\
 layer 1	&12.44\%	&18.67\% \\
 layer 2	&12.01\%	&16.30\% \\
 layer 3	&11.15\%	&15.07\% \\
 layer 4	&10.14\%	&13.95\% \\
 layer 5	&9.29\%	&12.24\% \\
 layer 6	&8.66\%	&10.99\% \\
 layer 7	&7.94\%	&9.00\% \\
 layer 8	&7.18\%	&7.78\% \\
 layer 9	&7.09\%	&7.70\% \\
 layer 10	&7.27\%	&8.18\% \\
 layer 11	&7.64\%	&8.79\% \\
 layer 12	&8.62\%	&9.24\% \\
 layer 13	&8.38\%	&9.12\% \\
 layer 14	&7.67\%	&8.73\% \\
 layer 15	& 6.95\%& \bf 7.79\% \\
 layer 16	&\bf 6.39\%	&8.11\% \\
 layer 17	&7.29\%	&9.05\% \\
 layer 18	&8.38\%	&11.45\% \\
 layer 19	&9.93\%	&14.81\% \\
 layer 20	&12.25\%	&20.76\% \\
 layer 21	&22.25\%	&30.43\% \\
 layer 22	&32.35\%	&37.83\% \\
 layer 23	&33.88\%	&38.70\% \\
final proj &	33.79\%	&38.89\% \\ \hline
\end{tabular}}
\caption{Per layer ABX scores for Wav2vec2.0 Base}
\label{fig:abx_small}
\end{table}


\end{document}